\documentclass[letterpaper, 10 pt, conference]{ieeeconf}  % Comment this line out if you need a4paper

\makeatletter
\let\NAT@parse\undefined
\makeatother

\IEEEoverridecommandlockouts                              % This command is only needed if 
\usepackage{graphicx} % for pdf, bitmapped graphics files
\usepackage[numbers]{natbib}
\usepackage{amsmath} % assumes amsmath package installed
\usepackage{amssymb}  % assumes amsmath package installed
\usepackage{mathtools}
\usepackage[]{algorithm2e}
\usepackage[caption=false]{subfig}
\usepackage{url}

\newcommand{\figref}[1]{Fig.~\ref{#1}}
\newcommand{\secref}[1]{Section~\ref{#1}}
\newcommand{\tabref}[1]{Table~\ref{#1}}

\newcommand{\norm}[1]{\left\lVert#1\right\rVert}

\title{\LARGE \bf
Continuous-time Radar-inertial Odometry for Automotive Radars
}

\author{Yin Zhi Ng$^{1,2}$, Benjamin Choi$^{1}$, Robby Tan$^{2}$ and Lionel Heng$^{1}$% <-this % stops a space
\thanks{$^{1}$Robotics Autonomy Lab, Robotics Division, DSO National Laboratories}%
\thanks{$^{2}$Department of Electrical and Computer Engineering, National University of Singapore}
}

\begin{document}

\maketitle
\thispagestyle{empty}
\pagestyle{empty}

\begin{abstract}
We present an approach for radar-inertial odometry which uses a continuous-time framework to fuse measurements from multiple automotive radars and an inertial measurement unit (IMU). Adverse weather conditions do not have a significant impact on the operating performance of radar sensors unlike that of camera and LiDAR sensors. Radar's robustness in such conditions and the increasing prevalence of radars on passenger vehicles motivate us to look at the use of radar for ego-motion estimation. A continuous-time trajectory representation is applied not only as a framework to enable heterogeneous and asynchronous multi-sensor fusion, but also, to facilitate efficient optimization by being able to compute poses and their derivatives in closed-form and at any given time along the trajectory. We compare our continuous-time estimates to those from a discrete-time radar-inertial odometry approach and show that our continuous-time method outperforms the discrete-time method. To the best of our knowledge, this is the first time a continuous-time framework has been applied to radar-inertial odometry.
\end{abstract}

%%%%%%%%%%%%%%%%%%%%%%%%%%%%%%%%%%%%%%%%%%%%%%%%%%%%%%%%%%%%%%%%%%%%%%%%%%%%%%%%
\section{INTRODUCTION}
Odometry involves the use of either proprioceptive or exteroceptive sensors to determine the ego-motion of the vehicle with respect to a local fixed frame. Accurate ego-motion estimation is important for closed-loop control of autonomous vehicles during navigation, and therefore, robust autonomy. The possibility of GPS outages and failures in map-based localization due to outdated maps causes a fallback to odometry, and accurate estimation of the autonomous vehicle's pose is required for the vehicle to safely navigate from one waypoint to another.

Popular forms of odometry involve the use of either cameras, LiDARs, or a combination of both, to estimate the vehicle's motion. State-of-the-art odometry methods incorporate IMU measurements to better predict motion and to eliminate roll and pitch drift. In this way, they can achieve ego-motion estimation with lower drift over long distances.  However, the performance of camera-based and LiDAR-based methods is heavily dependent on the quality of sensor data and starts to break down in adverse weather conditions. Most visual odometry approaches fail in low-light conditions due to the lack of texture in images, impeding feature detection, descriptor extraction or photometric matching. Moreover, in rainy conditions, the rain veiling effect can reduce the visibility range of cameras significantly, and rain streaks as well as raindrops adhered to camera lenses can degrade images considerably. While LiDAR odometry methods are typically resilient to low-light conditions, airborne obscurants resulting from adverse weather conditions such as fog, rain, and snow create noisy LiDAR returns and a veiling effect, both of which interfere with scan matching, and can reduce the accuracy of ego-motion estimation. 

Radar sensors utilize signals of a lower frequency along the electromagnetic spectrum compared to camera and LiDAR sensors. Due to the fact that signal attenuation due to atmospheric gases and rain decreases with decreasing frequency, radar sensors work better in adverse weather conditions \citep{hong2021radar,Kramer2021ISER}. This better performance coupled with their low cost are the main reasons why radar sensors are preferred in advanced driver assistance systems (ADAS) on passenger cars.

Improvements in automotive radar technology have seen automotive radars switching from the 24 GHz band to the 77 GHz band for better range accuracy and resolution. As a result of these improvements, new applications including radar-based odometry are being opened up beyond the original application of moving object tracking for adaptive cruise control. We can also take advantage of the fact that major car manufacturers have also begun including 77 GHz automotive radars as part of ADAS systems on their vehicles due to their low cost, long range and robustness in adverse weather conditions. Our radar-inertial odometry implementation can be easily deployed on such vehicles fitted with these ADAS systems, and does not require any vehicle modifications which can be costly.
%While radar odometry is not necessarily a replacement for methods involving the other sensors, it can be treated as an additional layer of redundancy during degraded operating conditions.

A continuous-time framework is increasingly popular in odometry approaches. It is advantageous as the smoothness of the vehicle's trajectory is enforced and enables robustness to occlusions faced by radar sensors. A continuous-time trajectory representation also enables us to compute poses and their derivatives at any given time and in closed-form, making it well-suited for asynchronous sensor fusion. Continuous-time frameworks have been applied to LiDAR-inertial odometry \citep{Kaul2016JFR} and visual-inertial odometry \citep{PatronPerez2015IJCV, Mueggler2018TRO}, but not radar-inertial odometry. To the best of our knowledge, this is the first time that a continuous-time framework has been applied to radar-inertial odometry.

Our contributions in this paper are:
\begin{enumerate}
\item Continuous-time trajectory representation for radar-inertial odometry,
\item Multiple asynchronous radars for radar-inertial odometry.
\end{enumerate}
A continuous-time trajectory representation comes with inherent smoothness constraints which are reasonable assumptions for odometry involving ground vehicles. During periods in which the radar sensor undergoes significant occlusions, measurements corresponding to static parts of the environment which are needed to estimate ego-motion can be too few, and cause degradation in the accuracy of radar odometry. Smoothness constraints inherent in a continuous-time trajectory can be leveraged to mitigate the degradation. In addition, as the number of radar sensors increases, the number of variables to be optimized does not increase, unlike those in discrete-time approaches, enabling our radar-inertial odometry to scale well to multiple radars.

We use multiple radars to minimize the likelihood of occlusion, which in turn, increases the number of measurements corresponding to static parts of the environment. Such measurements are used for ego-motion estimation; measurements corresponding to dynamic parts of the environment are considered as outliers and filtered out. Previous methods utilizing radar sensors either work with a single radar, or multiple synchronous radars. Our implementation does not require multiple radars to do synchronous capture, and instead, only requires radars to be time-synchronized with respect to a common clock.

\section{Related Work}

There are a few existing works on LiDAR-inertial odometry \citep{Kaul2016JFR, Ye2019ICRA, Shan2020IROS} and many more existing works on visual-inertial odometry \citep{Mourikis2007ICRA, Leutenegger2015IJRR, PatronPerez2015IJCV, Forster2017TRO, Mueggler2018TRO, Qin2018TRO}. We pay particular attention to those who use continuous-time frameworks. \citet{Kaul2016JFR} employ a continuous-time approach to LiDAR-inertial odometry for an UAV with a rotating payload consisting of a 2D LiDAR and an IMU. They use a cubic spline to represent the trajectory and optimise the trajectory by minimizing surfel-to-surfel distance errors based on LiDAR measurements and inertial errors. \citet{PatronPerez2015IJCV} use a continuous-time framework for visual-inertial odometry with a rolling-shutter camera. Similarly to \citep{Kaul2016JFR}, they optimize the cubic-spline-based trajectory by minimizing inertial errors and image reprojection errors instead of LiDAR-based correspondence errors. \citet{Mueggler2018TRO} apply the same continuous-time framework to event-based cameras.

Radar odometry techniques \citep{Cen2018ICRA, Park2020ICRA, Burnett2021RAL} exist for 2D imaging radar which outputs 2D radar images at low update rates. They use either feature-based methods or direct methods to estimate the motion between consecutive radar images. In contrast, automotive radars generate a list of target measurements at high update rates, each of which includes azimuth, optionally elevation, range, and radial velocity of the target with respect to the sensor. The ability of automotive radars to measure both range and radial velocity gives us the opportunity to estimate both the pose and velocity of the vehicle via correspondence-free methods. Such methods are advantageous as a scan from an automotive radar can be quite sparse, containing up to a few hundred target measurements and making radar scan matching difficult. With the prevalence of automotive radars on passenger vehicles, we choose to focus on the use of automotive radars for our work on radar-inertial odometry.

\citet{Kellner2013ITSC}'s seminal work on radar odometry proposes a least-squares solution to recover the radar sensor's velocity from multiple target measurements, and leverages the Ackermann steering model to estimate the vehicle's velocity. \citet{Kellner2014ICRA} extend their work to multiple synchronous radar sensors. We note that our work neither makes any assumption of the vehicle model nor requires multiple radar sensors to be synchronous. \citet{Kramer2020ICRA} propose a discrete-time approach to radar-inertial odometry. They estimate the vehicle's pose for each radar scan by solving a non-linear least-squares problem that comprises both radial velocity residuals and inertial residuals. We note that extending this discrete-time approach to multiple asynchronous radar sensors requires estimation of more vehicle poses within a given period of time due to the need to estimate the vehicle's pose for each radar scan. In contrast, a continuous-time approach only requires us to estimate the parameters of a trajectory representation whose dimensions are invariant to the number of radar sensors, and thus, is more computationally efficient. Other advantages of a continuous-time trajectory representation include smoothness constraints and the ability to infer the vehicle's pose and its derivatives at any point along the trajectory. Hence, we explore a continuous-time framework for our work on radar-inertial odometry.

\section{Notation}
We define the notations to be used throughout this paper. We denote the world reference frame as $\underrightarrow{\mathcal{F}}_w$. Without loss of generality, we assume the vehicle reference frame and IMU reference frame to coincide and denote the vehicle reference frame as $\underrightarrow{\mathcal{F}}_v$. The $x$-axis of $\underrightarrow{\mathcal{F}}_v$ lies along the longitudinal axis of the vehicle and points forward while the $y$-axis of $\underrightarrow{\mathcal{F}}_v$ lies along the lateral axis of the vehicle and points left. The radar sensor reference frame is denoted as $\underrightarrow{\mathcal{F}}_s$.

We denote the pose of the radar sensor with respect to $\underrightarrow{\mathcal{F}}_v$ as a rigid body transformation $\mathbf{T}_{v,s} \in SE(3)$ from $\underrightarrow{\mathcal{F}}_s$ to $\underrightarrow{\mathcal{F}}_v$ and whose rotation matrix part is $\mathbf{R}_{v,s}$ and translation part is $\mathbf{t}_{v,s}$. $\mathbf{T}_{v,s}$ can be obtained from extrinsic calibration. Likewise, we denote the pose of the radar sensor in $\underrightarrow{\mathcal{F}}_w$ as $\mathbf{T}_{w,s} = \mathbf{T}_{w,v} \, \mathbf{T}_{v,s}$ where $\mathbf{T}_{w,v}$ is the vehicle pose in $\underrightarrow{\mathcal{F}}_w$ to be estimated.

\section{Method}
\subsection{Continuous-time Trajectory Representation}
\begin{figure}
    \centering
    \includegraphics[width=1.0\columnwidth]{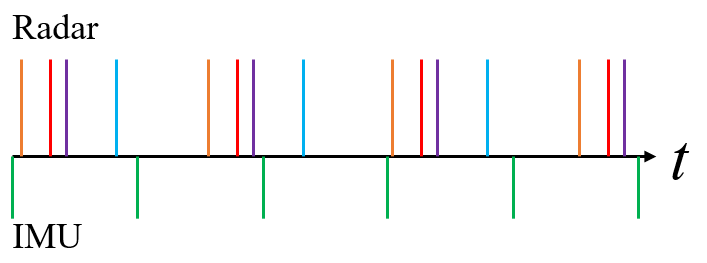}
    \caption{While IMU measurements are received at a constant rate, scans from multiple radars (represented by the different colours) do not arrive at consistent intervals and are rarely aligned with each other. A continuous-time trajectory representation is an ideal solution to asynchronous sensor fusion.}
    \label{fig:asynchronous}
\end{figure}

Discrete-time radar odometry methods typically make use of a single radar sensor for state estimation due to the asynchronous nature of the radar sensor. In instances involving the use of multiple radars \citep{Kellner2014ICRA}, they have to be synchronised with one another. Discrete-time radar-inertial odometry methods require radar scans to have corresponding IMU measurements. As it is rare for an IMU measurement to be captured at the same time as a radar scan, an interpolated IMU measurement has to be obtained instead. While continuous-time trajectory representations have been previously used as a way to reduce computational complexity, one key advantage of continuous-time trajectory representations is the ability to evaluate the trajectory at any given time, and which allows for fusion of data from asynchronous sensors \citep{Mueggler2018TRO}. This is ideal for our use case involving multiple radars and an IMU sensor, all of which are asynchronous. \figref{fig:asynchronous} gives a visual illustration of the timestamps of radar and IMU measurements as they get captured over time.

For the continuous-time trajectory representation, it is ideal to choose a minimum-degree spline that enforces a smooth trajectory, and allows for $C^2$ continuity and local support. $C^2$ continuity enables us to derive closed-form expressions for predicted IMU measurements and predicted radial velocity components of radar measurements, and local support facilitates trajectory optimization. Hence, we use the cubic B-spline, which is represented by four control poses parameterized in $SE(3)$; we follow the cubic B-spline parameterization of \citet{PatronPerez2015IJCV}.

\begin{figure}
    \centering
    \includegraphics[width=1.0\columnwidth]{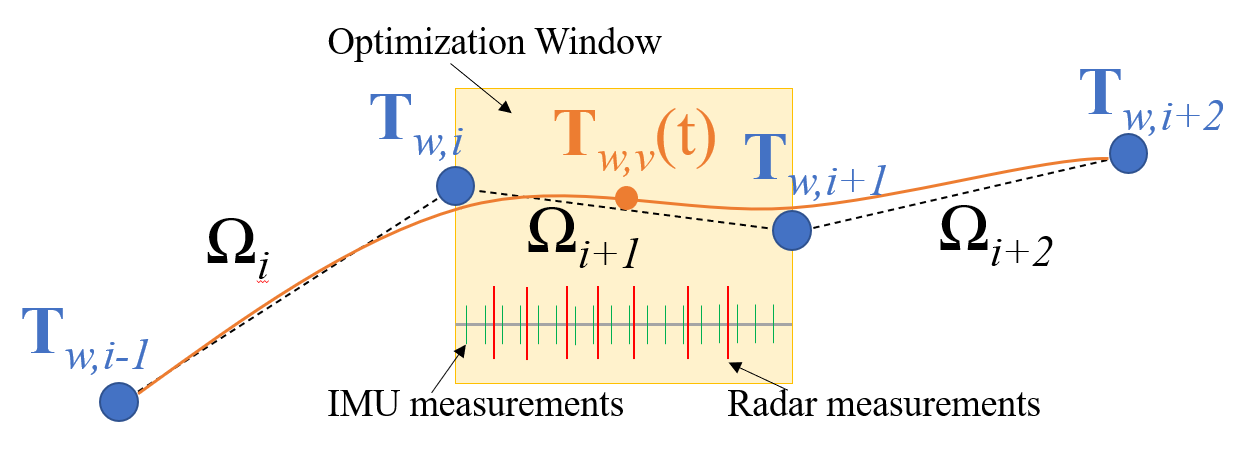}
    \caption{Cubic B-spline framework. Optimization is done over all IMU and radar measurements in the time window between $t_i$ and $t_{i+1}$.}
    \label{fig:Spline}
\end{figure}

For a given sensor measurement with timestamp $t$, we need to find the corresponding pose and its first and second derivatives in order to predict IMU and partial radar measurements, which are required for optimization. Before we can do so, we need to fit a spline trajectory through four control poses with timestamps $\{t_{i-1}, t_i, t_{i+1}, t_{i+2}\}$ for $t \in [t_{i},t_{i+1})$ and assuming that a control pose is created at the end of every uniform time interval $\Delta t$:
\begin{align}
    t_i &= i \times \Delta t \\
    u(t) &= \frac{t - t_i}{\Delta t},\ \ \ \ u(t) \in [0,1).
\end{align}
The pose at time $t$ along the spline trajectory is given by:
\begin{equation}
    \mathbf{T}_{w,v}(u(t)) = \mathbf{T}_{w,i-1} \prod_{j=1}^{3} 
    \exp(\Tilde{\mathbf{B}}_{j}(u(t))\Omega_{i+j-1}),
\end{equation}
where $\mathbf{T}_{w,i}$ is the control pose at time $t_i$, the incremental pose from $t_{i-1}$ to $t_i$ is derived from the twist
\begin{equation}
    \Omega_{i} = \log(\mathbf{T}_{w,i-1}^{-1}\mathbf{T}_{w,i}),
\end{equation}
and dropping the subscript $j$ without loss of generality and for ease of notation, the cumulative basis function $\Tilde{\mathbf{B}}(u)$, its first derivative $\dot{\Tilde{\mathbf{B}}}(u)$, and its second derivative $\ddot{\Tilde{\mathbf{B}}}(u)$, which can be computed with the matrix parameterization of the De Boor-Cox formula:
\begin{align}
    \Tilde{\mathbf{B}}(u) &= \mathbf{C}\begin{bmatrix}
    1 \\ u \\ u^{2} \\ u^{3} 
    \end{bmatrix}, \ \
    \mathbf{C} = \frac{1}{6}\begin{bmatrix}
    6 & 0 & 0 & 0 \\
    5 & 3 & {-3} & 1 \\
    1 & 3 & 3 & {-2} \\
    0 & 0 & 0 & 1
    \end{bmatrix}\\
    \dot{\Tilde{\mathbf{B}}}(u) &= \frac{1}{\Delta t}\mathbf{C}\begin{bmatrix}
    0 \\ 1 \\ 2u \\ 3u^{2}
    \end{bmatrix}, \ \
    \ddot{\Tilde{\mathbf{B}}}(u) = \frac{1}{\Delta t^{2}}\mathbf{C}\begin{bmatrix}
    0 \\ 0 \\ 2 \\ 6u
    \end{bmatrix}.
\end{align}
The first and second derivatives of the pose $\mathbf{T}_{w,v}(u)$ can be computed as:
\begin{align}
 \nonumber
   \dot{\mathbf{T}}_{w,v}(u) &= \mathbf{T}_{w,i-1}( \dot{\mathbf{A}}_{0}\mathbf{A}_{1}\mathbf{A}_{2}
   +\mathbf{A}_{0}\dot{\mathbf{A}}_{1}\mathbf{A}_{2}
   +\mathbf{A}_{0}\mathbf{A}_{1}\dot{\mathbf{A}}_{2}
   ), \\ \nonumber
   \ddot{\mathbf{T}}_{w,v}(u) &= \mathbf{T}_{w,i-1}( \ddot{\mathbf{A}}_{0}\mathbf{A}_{1}\mathbf{A}_{2}
   +\mathbf{A}_{0}\ddot{\mathbf{A}}_{1}\mathbf{A}_{2}
   +\mathbf{A}_{0}\mathbf{A}_{1}\ddot{\mathbf{A}}_{2}, \nonumber\\ \nonumber
   &\qquad +\dot{\mathbf{A}}_{0}\mathbf{A}_{1}\mathbf{A}_{2}
   +\mathbf{A}_{0}\dot{\mathbf{A}}_{1}\mathbf{A}_{2}
   +\mathbf{A}_{0}\mathbf{A}_{1}\dot{\mathbf{A}}_{2}
   ), \\ \nonumber
   \mathbf{A}_j &= \exp(\Omega_{i+j}\Tilde{\mathbf{B}}(u)_{j+1}), \\ \nonumber
   \dot{\mathbf{A}}_j &= \mathbf{A}_j\Omega_{i+j}\dot{\Tilde{\mathbf{B}}}(u)_{j+1}, \\ \nonumber
   \ddot{\mathbf{A}}_j &= \dot{\mathbf{A}}_j\Omega_{i+j}\dot{\Tilde{\mathbf{B}}}(u)_{j+1} + \mathbf{A}_j\Omega_{i+j}\ddot{\Tilde{\mathbf{B}}}(u)_{j+1}.
\end{align}
\figref{fig:Spline} gives a visual illustration of our cubic B-spline framework.

\subsection{Generative Model} \label{sec:gen_model}
With the given spline trajectory, we can synthesize the linear velocity $\boldsymbol{v}_v(u)$, angular velocity $\boldsymbol{\omega}_v(u)$, and linear acceleration $\mathbf{a}_v(u)$ in $\underrightarrow{\mathcal{F}}_v$, which are required for predicting IMU measurements and the radial velocity component of radar measurements:
\begin{align}
    \mathbf{v}_v(u) &= \mathbf{R}^{T}_{w,v}(u)\dot{\mathbf{t}}_{w,v}(u) \\
    \boldsymbol{\omega}_v(u) &= (\mathbf{R}^{T}_{w,v}(u)\dot{\mathbf{R}}_{w,v}(u))^\vee \\
    \mathbf{a}_v(u) &= \mathbf{R}^{T}_{w,v}(u)(\ddot{\mathbf{t}}_{w,v}(u) + \mathbf{g}_w)
\end{align}
where $\dot{\mathbf{R}}_{w,v}$, $\dot{\mathbf{t}}_{w,v}$ and  $\ddot{\mathbf{t}}_{w,v}$ are submatrices of $\dot{\mathbf{T}}_{w,v}$ and $\ddot{\mathbf{T}}_{w,v}$. $\mathbf{g}_w$ is the acceleration due to gravity in  $\underrightarrow{\mathcal{F}}_w$. $[.]^\vee$ is the vee operator that is the inverse of the skew operator.

\subsection{Propagation of Control Poses}
To generate the minimum number of four control poses for the cumulative cubic B-spline trajectory, we refer to \citep{Leutenegger2015IJRR} and its formulation of IMU kinematics, specifically the propagation of $\mathbf{t}_{w,v}$ and the rotation quaternion $\mathbf{q}_{w,v}$.
\begin{align}
    \dot{\mathbf{t}}_{w,v} &= \mathbf{v}_{w,v} \\
    \dot{\mathbf{q}}_{w,v} &= \frac{1}{2}\begin{bmatrix}
    -\frac{1}{2}\mathbf{\omega}_v \\
    0
    \end{bmatrix}^\oplus
\end{align}
where the $\oplus$ operator is a right-hand compound operator as defined in \citep{Barfoot2011Acta}. In practical applications, we only have access to IMU measurements between times $t_i$ to $t_{i+1}$. As such, we propagate from $\mathbf{T}_{w,i+1}$ to $\mathbf{T}_{w,i+2}$ by using the constant-velocity model:
\begin{align}
    \mathbf{T}_{w,i+2} = \mathbf{T}_{w,i+1}\mathbf{T}_{w,i}^{-1}\mathbf{T}_{w,i+1}.
\end{align}

\subsection{Radial Velocity Residuals for Radar Sensors}

\begin{figure}
  \centering
  \includegraphics[width=1.0\columnwidth]{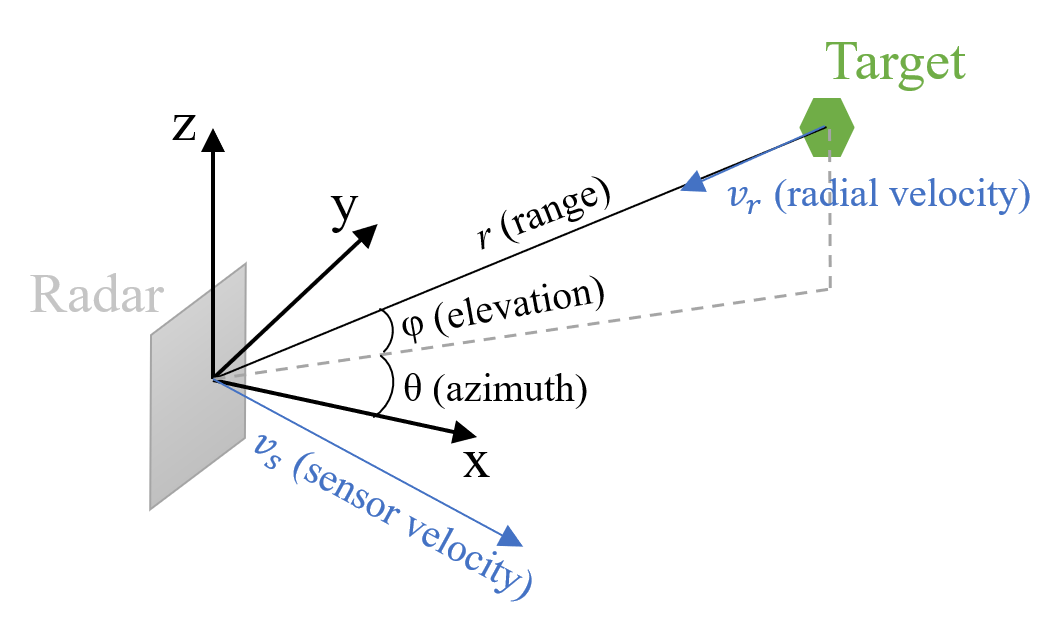}
  \caption{The radar sensor measures the radial component of a target's velocity relative to the radar sensor in addition to the target's range, azimuth, and elevation.}
  \label{fig:radar}
\end{figure}

\begin{figure}
  \centering
  \includegraphics[width=1.0\columnwidth]{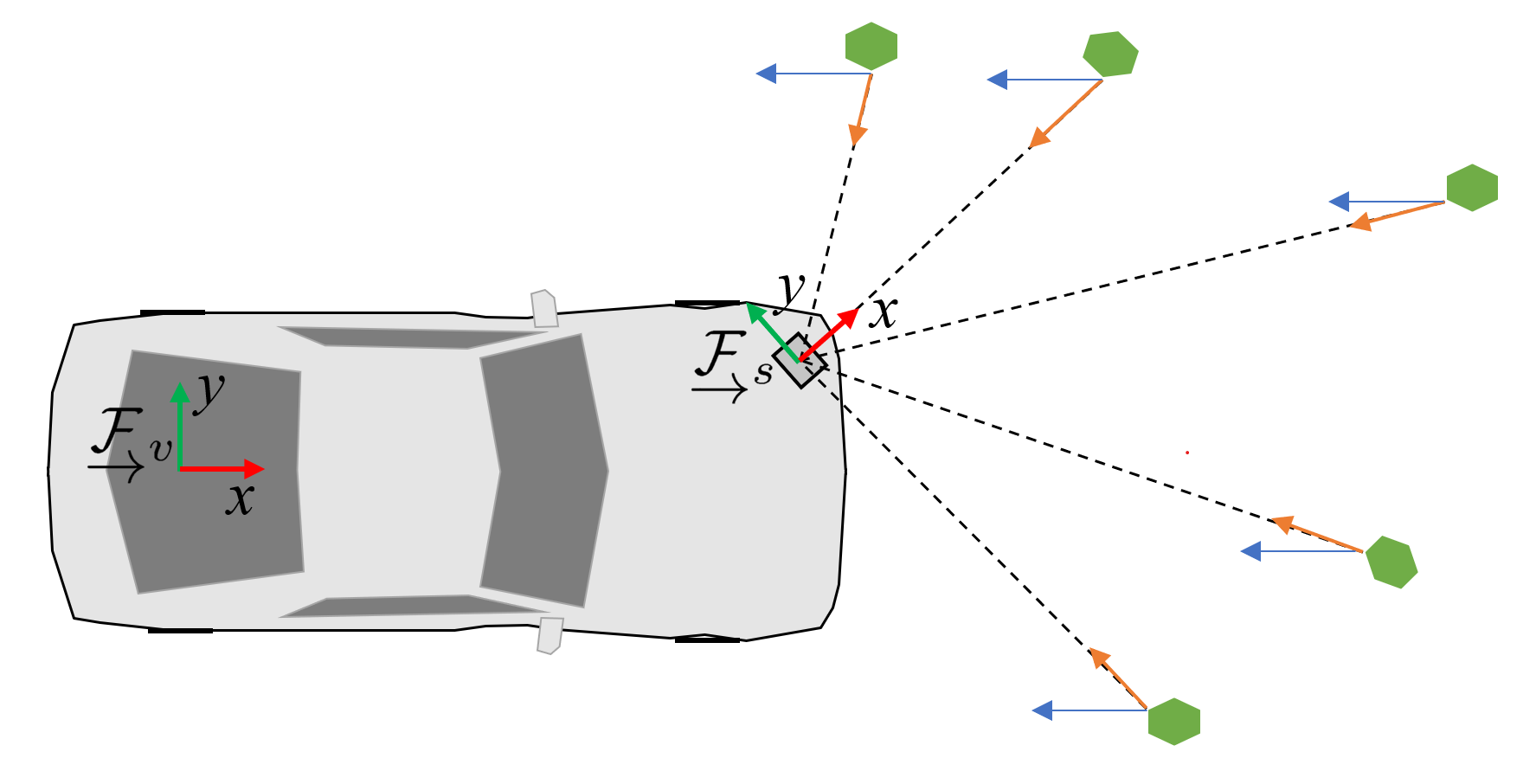}
  \caption{In this example, the vehicle is moving forward. Multiple stationary targets are observed by a radar sensor. For each radar target, a blue arrow indicates the velocity of the target with respect to the radar sensor. The orange arrow indicates the radial component of the target's velocity that is measured by the radar sensor.}
  \label{fig:vehicle_constrained_velocity}
\end{figure}

A single radar scan consists of $N$ target measurements. Each target measurement is represented by $[r, v_r, \theta, \phi]$ which corresponds to the range, radial velocity, azimuth, and elevation  respectively of the target with respect to the radar sensor. \figref{fig:radar} illustrates a target measurement with its attributes. The radial velocity of a static target as measured from the radar sensor is equal to the negation of the vector projection of the radar sensor's velocity in $\protect\underrightarrow{\mathcal{F}}_s$ and onto the unit direction vector from the radar sensor to the target:
\begin{align}
    v_r = -\mathbf{v}_s\cdot\begin{bmatrix}
    \cos\theta\cos\phi \\
    \sin\theta\cos\phi \\ 
    \sin\phi
    \end{bmatrix}
\end{align}
where $\mathbf{v}_s$ is the linear velocity of the sensor in $\underrightarrow{\mathcal{F}}_s$. We compute $\mathbf{v}_s$ using the following equation:
\begin{align}
    \mathbf{v}_s = \mathbf{R}_{s,v}(\mathbf{v}_v + \boldsymbol{\omega}_v\times\mathbf{t}_{v,s})
\end{align}
where $\mathbf{v}_v$ and $\boldsymbol{\omega}_v$ are derived from the generative model described in \secref{sec:gen_model}.
We compute the radial velocity residual $e_{r,n}$ for the $n$th target measurement in a radar scan:
\begin{align}
    e_{r,n} = \hat{v}_{r,n} - v_{r,n} \label{eq:radar_residual}
\end{align}
where $\hat{v}_{r,n}$ is the measured radial velocity associated with the target measurement.
\figref{fig:vehicle_constrained_velocity} illustrates the radial component of a target's velocity.
We note the assumption that the targets are stationary and only the vehicle is moving. However, in practice, there are moving objects in the environment in addition to measurement noise, and we apply a Cauchy loss function to the radar residual to reduce the influence of radar residuals corresponding to moving objects and measurement noise on the optimization which is described in the next section.

\subsection{Trajectory Optimization}
We formulate our continuous-time radar-inertial odometry as a non-linear least-squares problem for which we need to find the spline trajectory parameters that minimizes the following cost function:
\begin{align}
    F &= \frac{1}{K}\sum^K_{k=1}\frac{1}{N_k}\sum^{N_k}_{n=1}\frac{1}{\sigma_{v_r}^2}\rho(e_{r,n}(u(t_k)))^2 \nonumber \\
      &\qquad + \frac{1}{M}\sum^M_{j=1}\frac{1}{\sigma_\omega^2}(\hat{\boldsymbol{\omega}}_j-\boldsymbol{\omega}(u(t_j)) - \mathbf{b}_{\omega})^2 \nonumber \\
      &\qquad + \frac{1}{M}\sum^M_{j=1}\frac{1}{\sigma_a^2}(\hat{\boldsymbol{a}}_j-\boldsymbol{a}(u(t_j)) - \mathbf{b}_{a})^2  \nonumber \\
      &\qquad + \norm{\mathbf{e}_{b_\omega}}^2 + \norm{\mathbf{e}_{b_a}}^2.
\end{align}
For the first term, $K$ is the number of radar scans between times $t_i$ and $t_{i+1}$, $N_k$ is the number of targets for each radar scan $k$, $\rho$ is a loss function that reduces the influence of outliers on the solution to the least-squares problem, $\sigma_{v_r}$ is the radial velocity measurement noise parameter, and $e_r$ is the radial velocity residual from equation \eqref{eq:radar_residual}. The next two terms define the inertial residuals with $\hat{\boldsymbol{\omega}}_j$ and $\hat{\mathbf{a}}_j$ corresponding to gyroscope and accelerometer measurements respectively. $M$ is the total number of inertial measurements between $\mathbf{T}_{w,i}$ and $\mathbf{T}_{w,i+1}$, and $\sigma_\omega$ and $\sigma_a$ are the white noise parameters associated with the IMU. $\mathbf{b}_{\omega}$ and $\mathbf{b}_a$ are the gyroscope and accelerometer biases respectively. We define the bias errors $\mathbf{e}_{b_\omega}$ and $\mathbf{e}_{b_a}$ as:
\begin{align}
    \mathbf{e}_{b_\omega} &= \frac{1}{\sigma_{bg}}(\mathbf{b}_{\omega,i} - \mathbf{b}_{\omega,i+1}) \\
    \mathbf{e}_{b_a} &= \frac{1}{\sigma_{ba}}(\mathbf{b}_{a,i} - \mathbf{b}_{a,i+1})
\end{align}
whereby the biases $\mathbf{b}_{\omega,i}$ and $\mathbf{b}_{a,i}$ correspond to control pose $\mathbf{T}_{w,i}$, the biases $\mathbf{b}_{\omega,i+1}$ and $\mathbf{b}_{a,i+1}$ correspond to control pose $\mathbf{T}_{w,i+1}$, and $\sigma_{bg}$ and $\sigma_{ba}$ are the random walk parameters associated with the IMU.
We solve this optimization problem using the Google Ceres Solver \citep{ceres-solver}, an open-source library used for solving non-linear least-squares problems.

\section{EXPERIMENTS AND RESULTS}

\subsection{Experimental Setup}

\begin{figure}
  \centering
  \includegraphics[width=1.0\columnwidth]{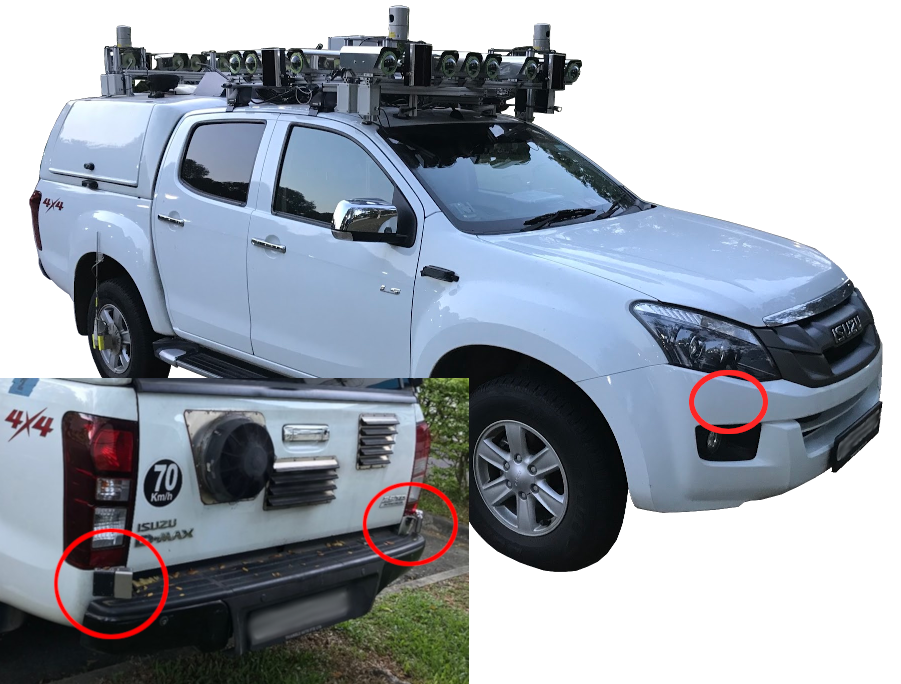}
  \caption{The vehicle platform. Red circles mark the locations of a radar located behind the front bumper and two radars mounted on the rear bumper.}
  \label{fig:vehicle_w_radars}
\end{figure}

An Isuzu D-Max vehicle as shown in \figref{fig:vehicle_w_radars} is used as the vehicle platform for our experiments. A dual-antenna GNSS/INS system with a tactical-grade IMU is installed in the vehicle, and provides both IMU data and ground-truth pose and velocity measurements at 200 Hz. The reference frame of the GNSS/INS system coincides with those of the vehicle and IMU, i.e. we use $\underrightarrow{\mathcal{F}}_v$ to refer to any of these three reference frames. Four smartmicro UMRR-8F T146 automotive radars are installed in an all-surround configuration: two behind the front bumper, and two on the rear bumper. Each automotive radar provides a scan containing up to 255 target measurements at 20 Hz. For each target measurement, the range accuracy is the larger of 0.5 m and 1\% of the measured range, the azimuth accuracy is $1^\circ$, and the elevation accuracy is $2^\circ$. The automotive radars are set to long-range mode for use in everyday operating conditions. All sensor data is hardware-timestamped to sub-millisecond precision, which is made possible through the use of a time server. Extrinsic calibration of the automotive radars is performed using the automatic targetless calibration method described in \citep{Heng2020IROS}.

\subsection{Experimental Evaluation}
We drive the vehicle along a route in an urban environment, and record sensor data from the start to the end of the route. The vehicle starts on top of a hill, goes through a heavily built-up area, before going on a highway and returning to the starting point. The route length is 5.8 km, and the vehicle reaches a maximum speed of 63 km/h. In general, the route contains a number of turns and altitude changes. There are many moving vehicles and pedestrians along the route, reducing the number of static target measurements available for ego-motion estimation. This makes the recorded dataset ideal for evaluating our approach.

We compare our continuous-time approach with a discrete-time radar-inertial odometry implementation based on the approach described in \citep{Kramer2020ICRA}. There are two key differences between our implementation and \citet{Kramer2020ICRA}'s implementation: (1) they assume the reference frames of the radar sensor and IMU to coincide whereas we do not, and (2) they only estimate the attitude, linear velocity and IMU biases whereas we include the position in addition to all these. In our discrete-time implementation, we formulate the 15-dimensional state vector to include the position, attitude, linear velocity and IMU biases, following the convention of  \cite{Leutenegger2015IJRR}.

We use two evaluation metrics to quantify the accuracy of our continuous-time approach. We use post-processed GNSS/INS data as ground truth. As used in \citep{Kramer2020ICRA}, the first metric is the the root-mean-squared error (RMSE) of the velocity and attitude estimates from our continuous-time approach estimates.
%\begin{equation}
%    RMSE = \sqrt{\frac{\sum^N_{i=1}(est - gt)^2}{N}}
%\end{equation}
%where $N$ refers to the total number of samples.
For the second metric, we use the metric used by the KITTI odometry benchmark \citep{Geiger2012CVPR} to compute the translational and rotational errors. In particular, we compute these errors over all possible sequences of length $\{100, 200, ... , 800\}$ meters, and average errors over all sequences. We compute the translational errors in both 2D and 3D.

\begin{figure}
    \centering
    \includegraphics[width=1.0\columnwidth, trim={3cm 0cm 3cm 0cm}]{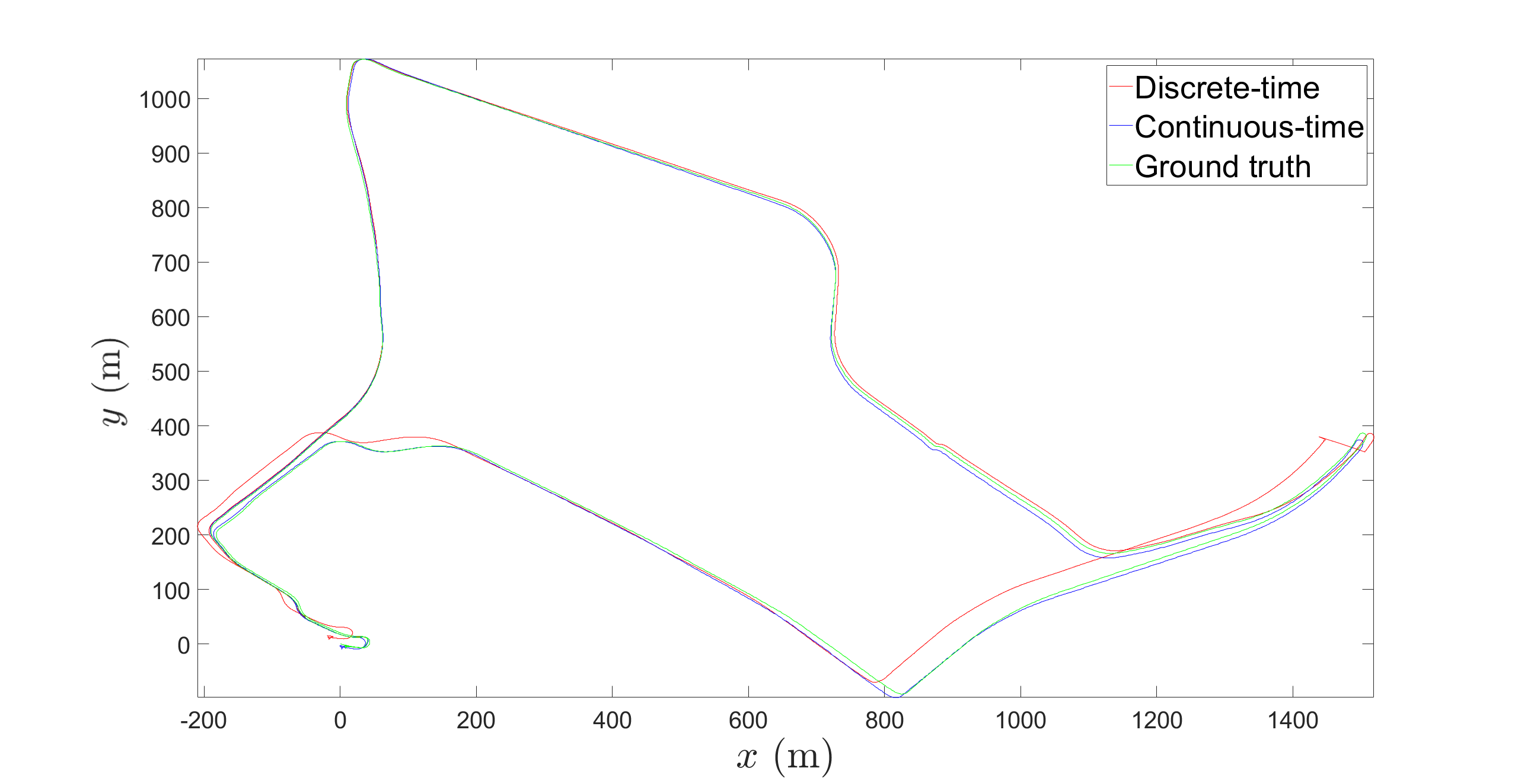}
    \caption{2D trajectory plots. Green represents ground truth, blue for continuous-time, and red for discrete-time.}
    \label{fig:2d}
\end{figure}
\begin{figure}
    \centering
    \includegraphics[width=1.0\columnwidth, trim={3cm 0cm 3cm 0cm}]{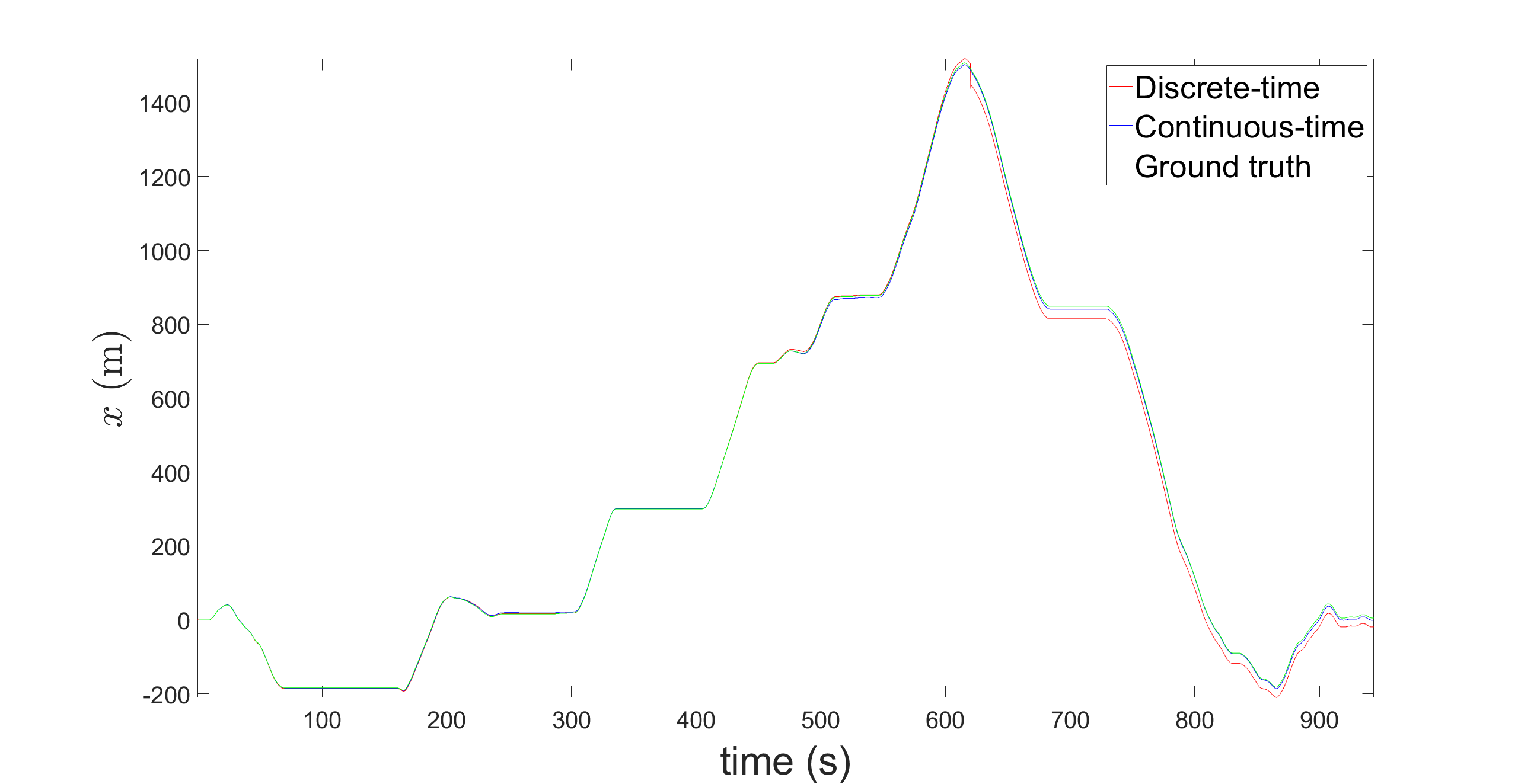}
    \includegraphics[width=1.0\columnwidth, trim={3cm 0cm 3cm 0cm}]{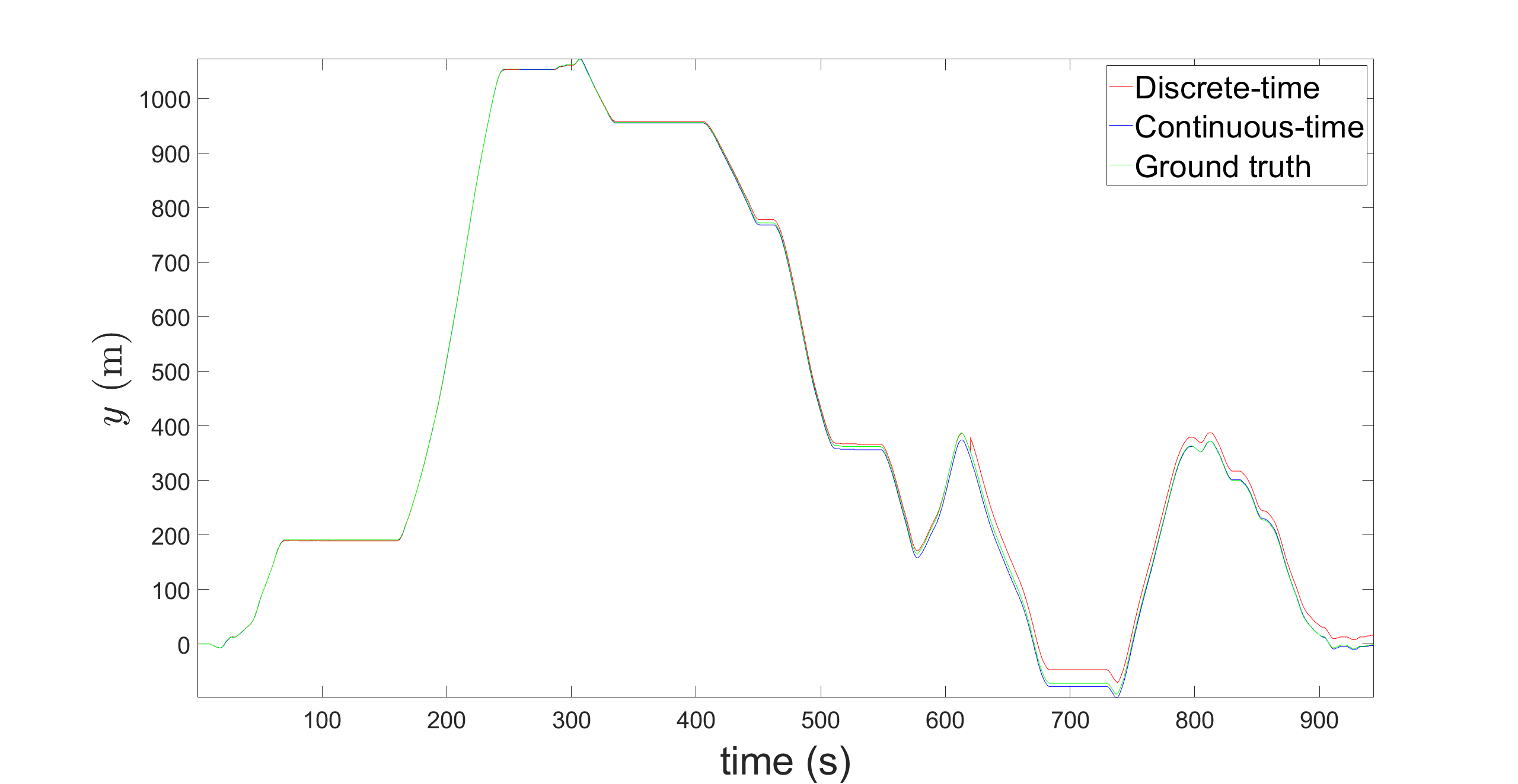}
    \includegraphics[width=1.0\columnwidth, trim={3cm 0cm 3cm 0cm}]{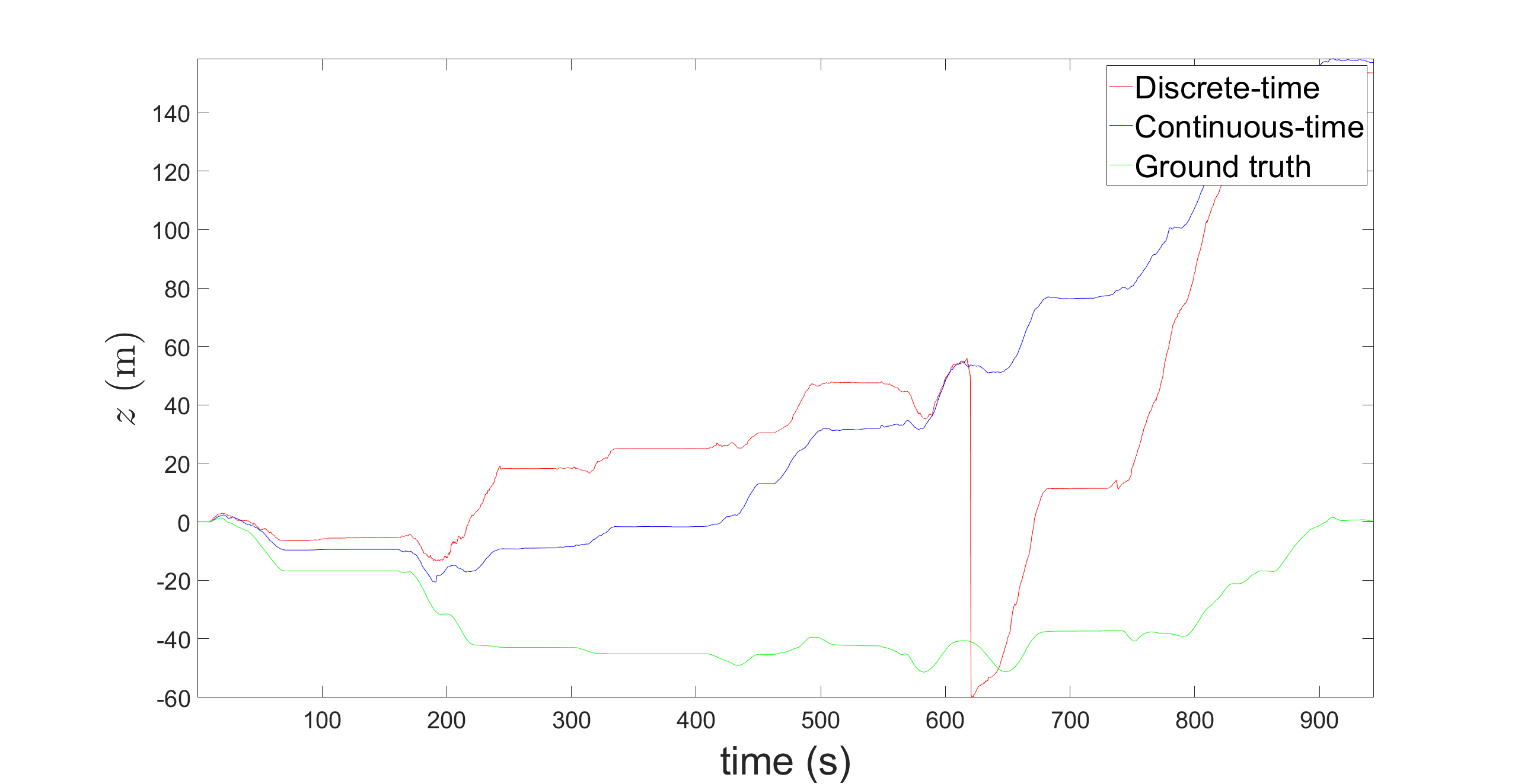}
    \caption{Trajectory estimates in each Cartesian plane against time. Green represents ground truth, blue for continuous-time, and red for discrete-time.}
    \label{fig:trajectory}
\end{figure}

\begin{figure}
    \centering
    \includegraphics[width=1.0\columnwidth, trim={3cm 0cm 3cm 0cm}]{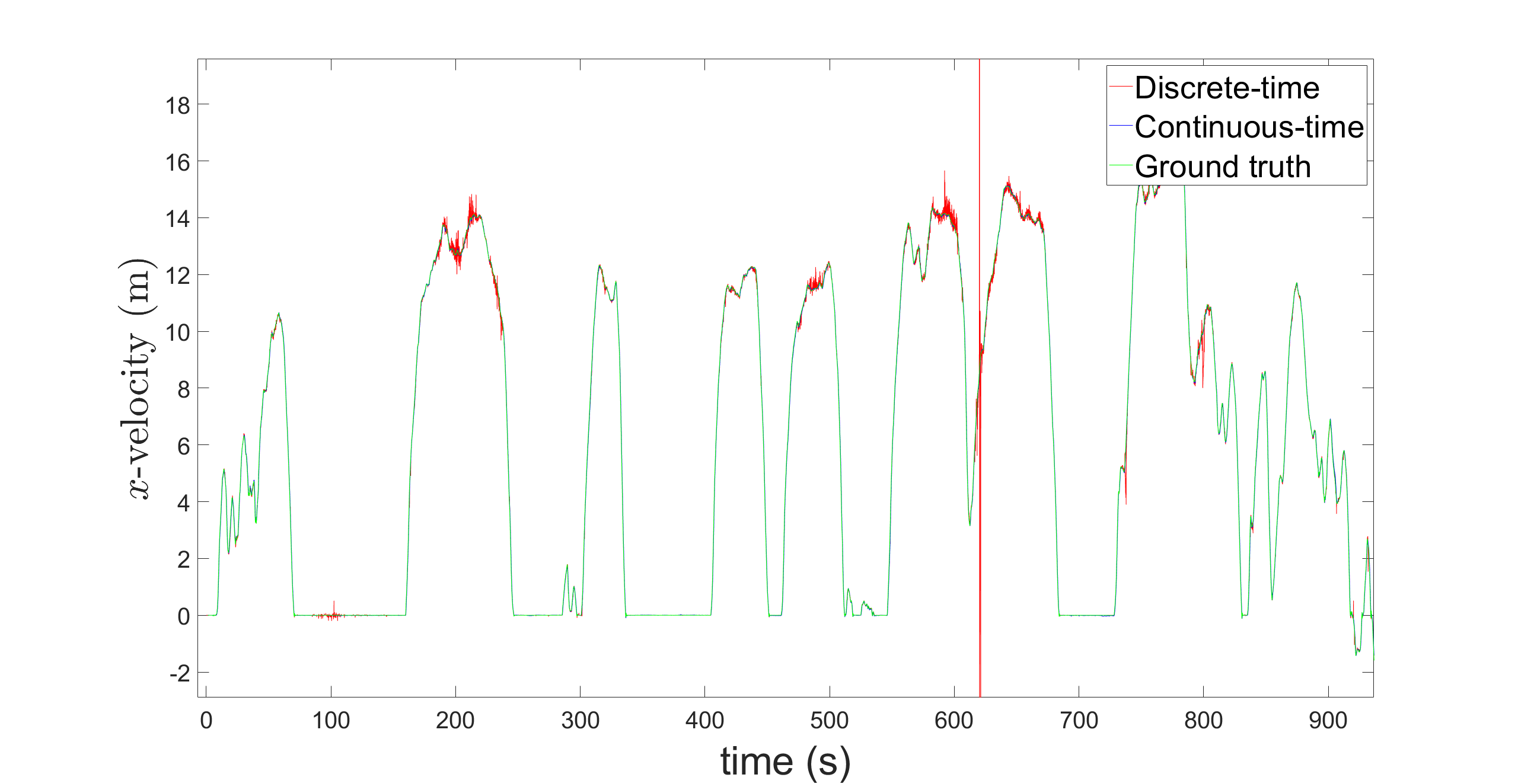}
    \includegraphics[width=1.0\columnwidth, trim={3cm 0cm 3cm 0cm}]{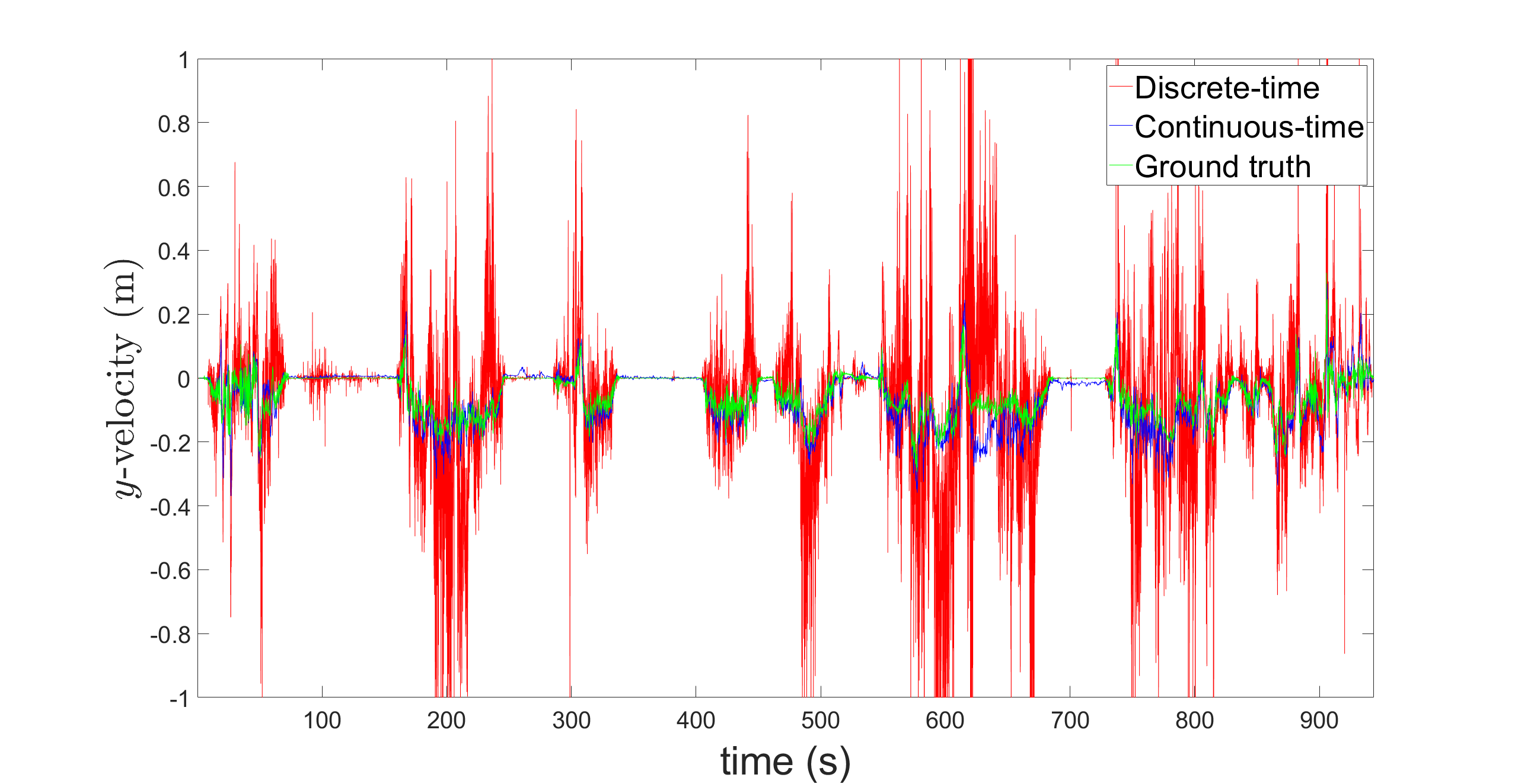}
    \includegraphics[width=1.0\columnwidth, trim={3cm 0cm 3cm 0cm}]{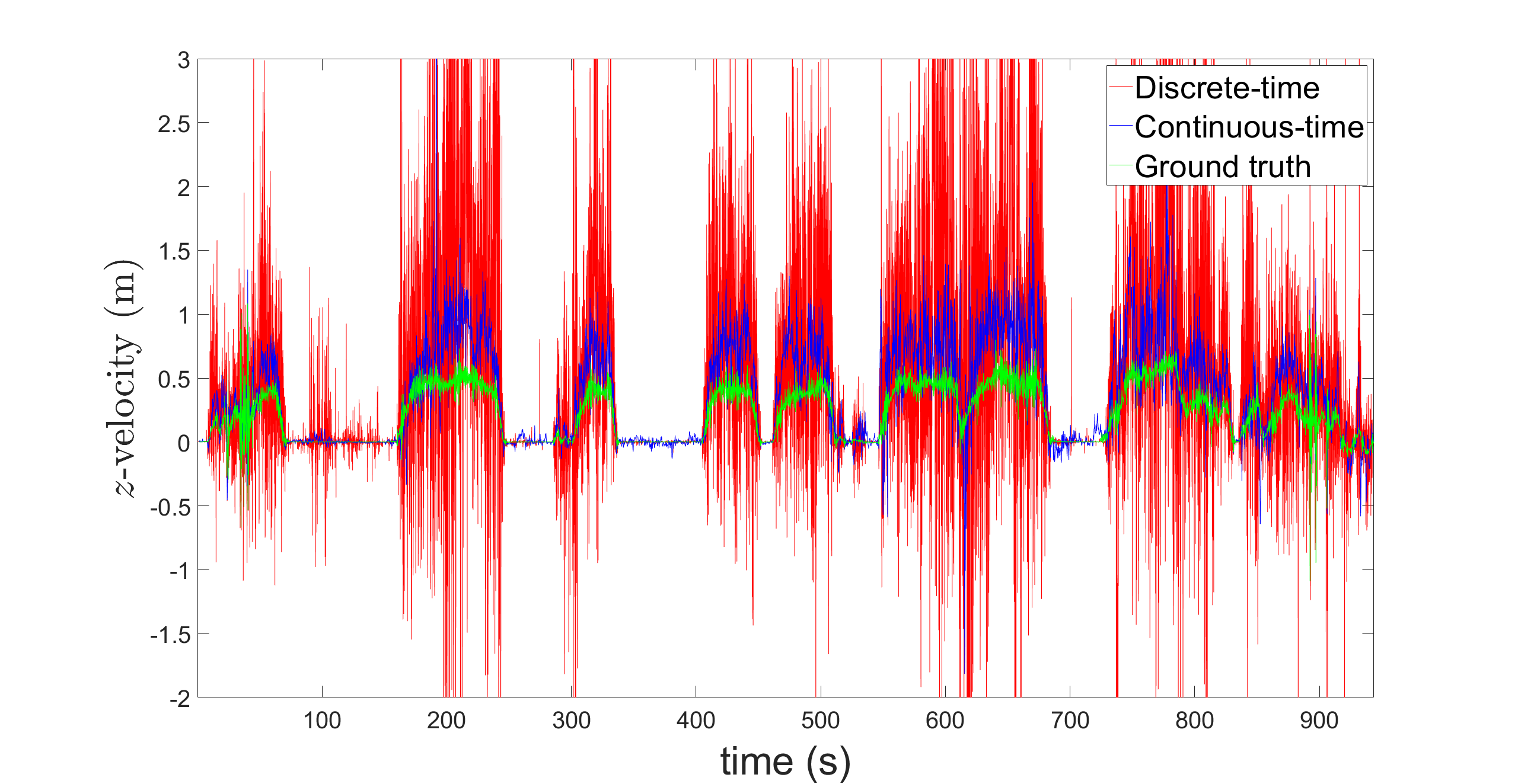}
    \caption{Velocity estimates in $\protect\underrightarrow{\mathcal{F}}_v$ in each Cartesian plane against time. Green represents ground truth, blue for continuous-time, and red for discrete-time.}
    \label{fig:velocity}
\end{figure}

\begin{figure}
    \centering
    \includegraphics[width=1.0\columnwidth, trim={3cm 0cm 3cm 0cm}]{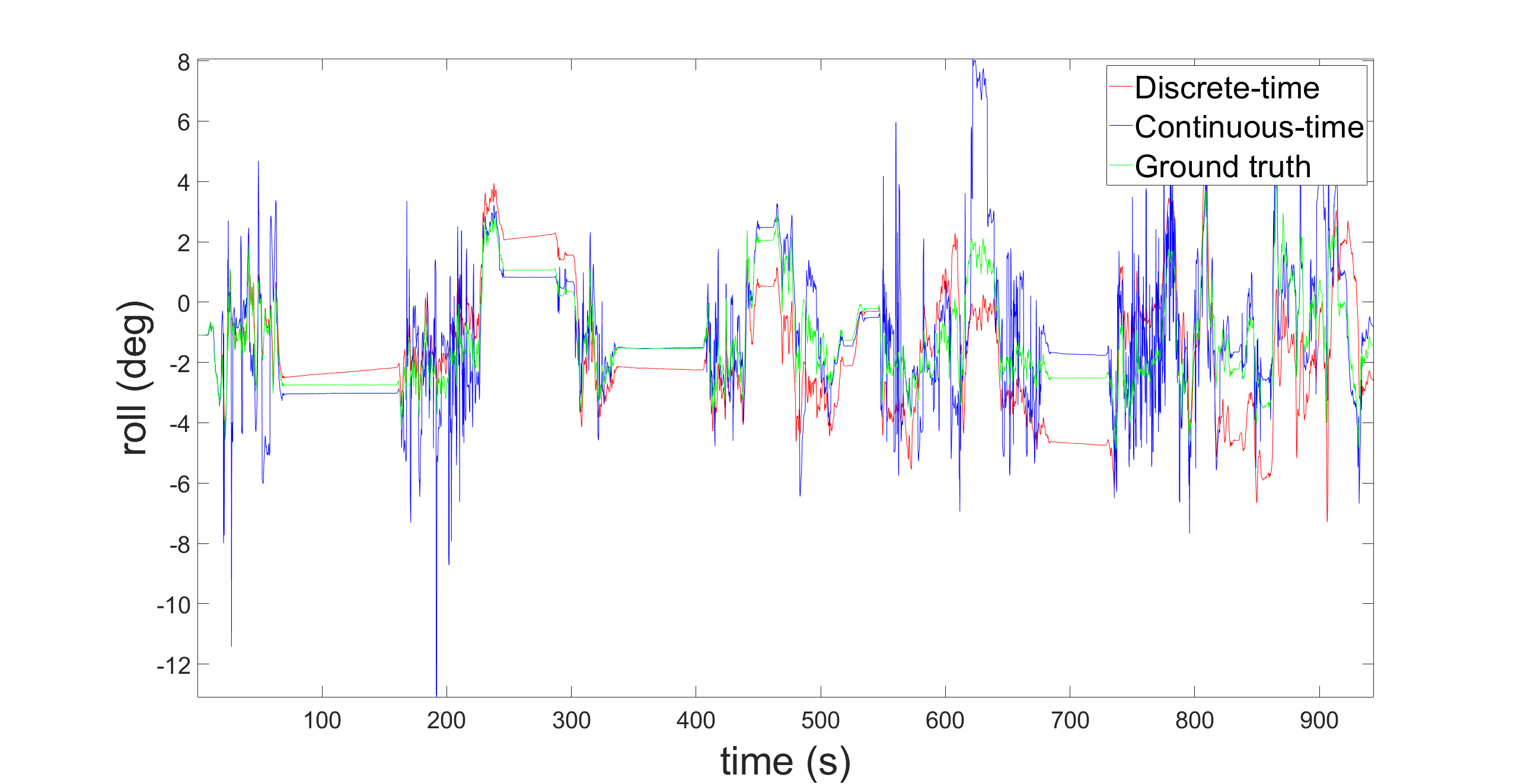}
    \includegraphics[width=1.0\columnwidth, trim={3cm 0cm 3cm 0cm}]{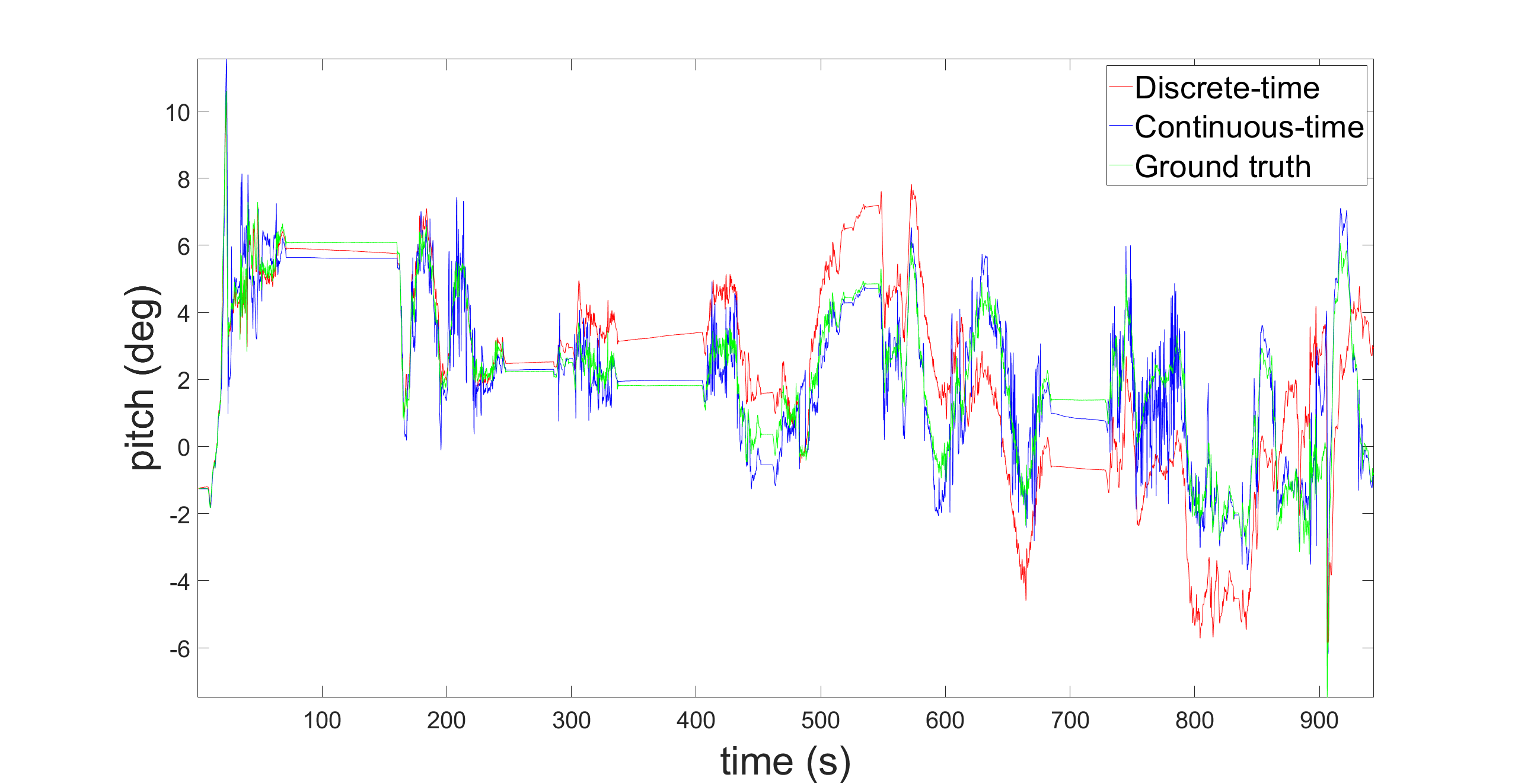}
    \includegraphics[width=1.0\columnwidth, trim={3cm 0cm 3cm 0cm}]{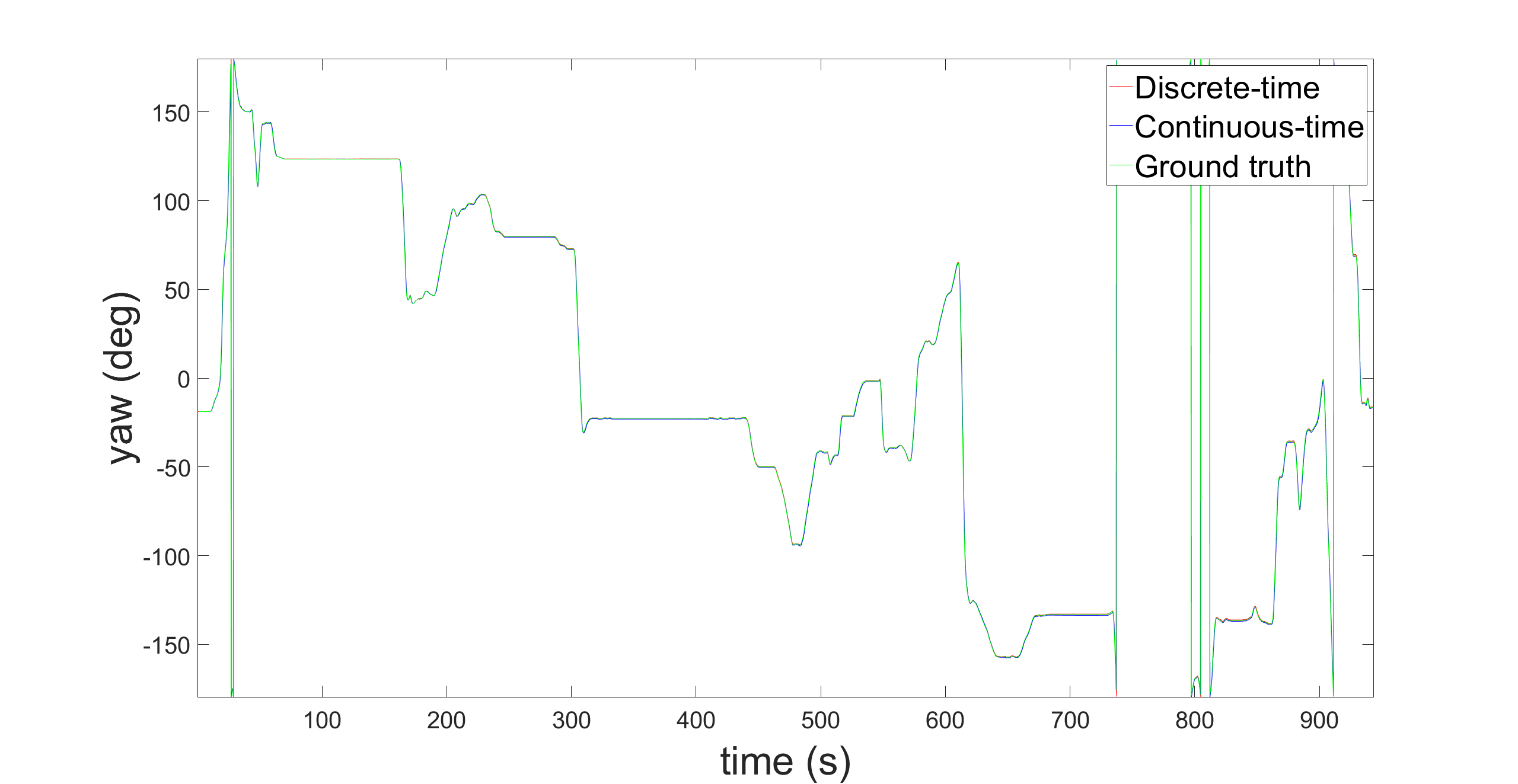}
    \caption{Roll, pitch, and yaw estimates against time. Green represents ground truth, blue for continuous-time, and red for discrete-time.}
    \label{fig:rpy}
\end{figure}

\subsection{Results}
For all plots, we use the green, red, and blue colours to represent estimates corresponding to the ground truth, the discrete-time approach, and our continuous-time approach. \figref{fig:2d} shows the estimated $x-y$ trajectory while \figref{fig:trajectory} plots each component of the estimated translation over time. We observe that the $x$ and $y$ components of the translation estimated by our continuous-time approach are closer to the ground truth compared to the discrete-time approach. We also observe that the $z$-translation is generally not accurate for both the continuous-time and discrete-time approaches as the radial velocity measurements are dominant in the $x$ and $y$ directions and do not sufficiently constrain the $z$-translation estimates. However, we note that the $z$-translation estimated by our continuous-time approach is not subjected to large magnitude shifts over short periods of time, demonstrating the benefits of $C^2$ continuity. 

\figref{fig:velocity} plots each component of the estimated velocity with respect to $\underrightarrow{\mathcal{F}}_v$ and over time. The velocity estimates from our continuous-time approach are much closer to the ground truth than those from the discrete-time approach. The velocity estimates from the discrete-time approach have significant errors due to the lack of sufficient static target measurements at many time instances. This reiterates the need for multiple radars and $C^2$ continuity for more robust odometry.

\figref{fig:rpy} plots each component of the estimated attitude over time. We observe that the roll and pitch estimates from our continuous-time approach are closer to the ground-truth compared to those from the discrete-time approach. One intuition behind why our continuous-time approach estimates roll and pitch more accurately is that the continuous-time trajectory representation allows us to estimate the gyroscope and accelerometer biases more accurately in the presence of significant radar measurement noise as the trajectory smoothness is enforced.

We tabulate the RMSE errors for each component of the velocity and attitude estimated by our continuous-time approach and the discrete-time approach in \tabref{tab:rmse}. We also tabulate the translational and rotational errors based on the KITTI odometry benchmark \citep{Geiger2012CVPR} and in \tabref{tab:kitti}. It is clear from both tables that our continuous-time approach significantly outperforms the discrete-time approach. In addition, we observe that the 2D translational error associated with our continuous-time approach is significantly lower than the reported 2D translational error of 3.58\% for a state-of-the-art radar odometry approach \citep{Burnett2021RAL} based on 2D imaging radar.

\begin{table}[]
\caption{Our continuous-time approach vs the discrete-time approach based on RMSE error metrics} \label{tab:rmse} 
\begin{center}
\begin{tabular}{|l|c|c|}
\hline
                         & Continuous-time & Discrete-time \\ \hline
$x$-velocity Error (m/s) & 0.1132          & 1.263 \\ \hline
$y$-velocity Error (m/s) & 0.0488          & 4.9317 \\ \hline
$z$-velocity Error (m/s) & 0.3173          & 4.5853 \\ \hline
Roll Error (deg)         & 1.4946          & 1.4569 \\ \hline
Pitch Error (deg)        & 0.8032           & 1.8107 \\ \hline
Yaw Error (deg)          & 1.3769          & 2.3092 \\ \hline
\end{tabular}
\end{center}
\end{table}

\begin{table}[]
\caption{Our continuous-time approach vs the discrete-time approach based on KITTI odometry error metrics} \label{tab:kitti} 
\begin{center}
\begin{tabular}{|l|c|c|}
\hline
           & Continuous-time & Discrete-time \\ \hline
2D Translational Error (\%) & 1.0645 & 1.8642 \\ \hline
3D Translational Error (\%) & 3.1762 & 5.3569 \\ \hline
Rotational Error (deg/m) & 0.0082 & 0.0036 \\ \hline
\end{tabular}
\end{center}
\end{table}

\subsection{Run-time Performance}
We run our implementation on a computer with an Intel Core i7-1165G7 2.80 GHz CPU. At the end of every 0.2-second time interval, our implementation creates and propagates a new control pose, and uses a 0.6-second sliding window of measurements for optimization. This iteration takes an average of 70 milliseconds on average, showing that our implementation is able to run faster than real-time.

\section{CONCLUSIONS}
In this paper, we have presented a radar-inertial odometry method utilizing a continuous-time framework. Our approach is able to fuse data from sensors that are asynchronous in nature. With the use of cubic B-splines as our continuous-time trajectory representation, we also enforce smoothness on the vehicle's trajectory. Previous radar odometry methods have proven the advantages of using radar sensors for odometry in challenging operating conditions. Our continuous-time method builds on those key principles and significantly improves odometry accuracy, outperforming the discrete-time approach while providing the flexibility for fusing data from other asynchronous sensors.

% %%%%%%%%%%%%%%%%%%%%%%%%%%%%%%%%%%%%%%%%%%%%%%%%%%%%%%%%%%%%%%%%%%%%%%%%%%%%%%%%

\bibliographystyle{abbrvnat}
\bibliography{references}

\begin{thebibliography}{21}
\providecommand{\natexlab}[1]{#1}
\providecommand{\url}[1]{\texttt{#1}}
\expandafter\ifx\csname urlstyle\endcsname\relax
  \providecommand{\doi}[1]{doi: #1}\else
  \providecommand{\doi}{doi: \begingroup \urlstyle{rm}\Url}\fi

\bibitem[Agarwal et~al.()Agarwal, Mierle, and Others]{ceres-solver}
S.~Agarwal, K.~Mierle, and Others.
\newblock Ceres solver.
\newblock \url{http://ceres-solver.org}.

\bibitem[Barfoot et~al.(2011)Barfoot, Forbes, and Furgale]{Barfoot2011Acta}
T.~Barfoot, J.~Forbes, and P.~Furgale.
\newblock Pose estimation using linearized rotations and quaternion algebra.
\newblock \emph{Acta Astronautica}, 68\penalty0 (1-2):\penalty0 101--112, 2011.

\bibitem[Burnett et~al.(2021)Burnett, Schoellig, and Barfoot]{Burnett2021RAL}
K.~Burnett, A.~Schoellig, and T.~Barfoot.
\newblock Do we need to compensate for motion distortion and doppler effects in
  spinning radar navigation?
\newblock \emph{IEEE Robotics and Automation Letters (RA-L)}, 6\penalty0
  (2):\penalty0 771--778, 2021.

\bibitem[Cen and Newman(2018)]{Cen2018ICRA}
S.~Cen and P.~Newman.
\newblock Precise ego-motion estimation with millimeter-wave radar under
  diverse and challenging conditions.
\newblock In \emph{IEEE International Conference on Robotics and Automation
  (ICRA)}, 2018.

\bibitem[Forster et~al.(2017)Forster, Carlone, Dellaert, and
  Scaramuzza]{Forster2017TRO}
C.~Forster, L.~Carlone, F.~Dellaert, and D.~Scaramuzza.
\newblock On-manifold preintegration for real-time visual--inertial odometry.
\newblock \emph{IEEE Transactions on Robotics (T-RO)}, 33\penalty0
  (1):\penalty0 1--21, 2017.

\bibitem[Geiger et~al.(2012)Geiger, Lenz, and Urtasun]{Geiger2012CVPR}
A.~Geiger, P.~Lenz, and R.~Urtasun.
\newblock Are we ready for autonomous driving? the kitti vision benchmark
  suite.
\newblock In \emph{IEEE Conference on Computer Vision and Pattern Recognition
  (CVPR)}, 2012.

\bibitem[Heng(2020)]{Heng2020IROS}
L.~Heng.
\newblock Automatic targetless extrinsic calibration of multiple 3d lidars and
  radars.
\newblock In \emph{IEEE/RSJ International Conference on Intelligent Robots and
  Systems (IROS)}, 2020.

\bibitem[Hong et~al.(2021)Hong, Petillot, Wallace, and Wang]{hong2021radar}
Z.~Hong, Y.~Petillot, A.~Wallace, and S.~Wang.
\newblock Radar slam: A robust slam system for all weather conditions, 2021.

\bibitem[Kaul et~al.(2016)Kaul, Zlot, and Bosse]{Kaul2016JFR}
L.~Kaul, R.~Zlot, and M.~Bosse.
\newblock Continuous-time three-dimensional mapping for micro aerial vehicles
  with a passively actuated rotating laser scanner.
\newblock \emph{Journal of Field Robotics (JFR)}, 33\penalty0 (1):\penalty0
  103--132, 2016.

\bibitem[Kellner et~al.(2013)Kellner, Barjenbruch, Klappstein, Dickmann, and
  Dietmayer]{Kellner2013ITSC}
D.~Kellner, M.~Barjenbruch, J.~Klappstein, J.~Dickmann, and K.~Dietmayer.
\newblock Instantaneous ego-motion estimation using doppler radar.
\newblock In \emph{IEEE International Conference on Intelligent Transportation
  Systems (ITSC)}, 2013.

\bibitem[Kellner et~al.(2014)Kellner, Barjenbruch, Klappstein, Dickmann, and
  Dietmayer]{Kellner2014ICRA}
D.~Kellner, M.~Barjenbruch, J.~Klappstein, J.~Dickmann, and K.~Dietmayer.
\newblock Instantaneous ego-motion estimation using multiple doppler radars.
\newblock In \emph{IEEE International Conference on Robotics and Automation
  (ICRA)}, 2014.

\bibitem[Kramer and Heckman(2020)]{Kramer2021ISER}
A.~Kramer and C.~Heckman.
\newblock Radar-inertial state estimation and obstacle detection for
  micro-aerial vehicles in dense fog.
\newblock In \emph{International Symposium on Experimental Robotics (ISER)},
  2020.

\bibitem[Kramer et~al.(2020)Kramer, Stahoviak, Santamaria-Navarro, akbar
  Agha-mohammadi, and Heckman]{Kramer2020ICRA}
A.~Kramer, C.~Stahoviak, A.~Santamaria-Navarro, A.~akbar Agha-mohammadi, and
  C.~Heckman.
\newblock Radar-inertial ego-velocity estimation for visually degraded
  environments.
\newblock In \emph{IEEE International Conference on Robotics and Automation
  (ICRA)}, 2020.

\bibitem[Leutenegger et~al.(2015)Leutenegger, Lynen, Bosse, Siegwart, and
  Furgale]{Leutenegger2015IJRR}
S.~Leutenegger, S.~Lynen, M.~Bosse, R.~Siegwart, and P.~Furgale.
\newblock Keyframe-based visual–inertial odometry using nonlinear
  optimization.
\newblock \emph{International Journal of Robotics Research (IJRR)}, 34\penalty0
  (3):\penalty0 314--334, 2015.

\bibitem[Mourikis and Roumeliotis(2007)]{Mourikis2007ICRA}
A.~Mourikis and S.~Roumeliotis.
\newblock A multi-state constraint kalman filter for vision-aided inertial
  navigation.
\newblock In \emph{IEEE International Conference on Robotics and Automation
  (ICRA)}, 2007.

\bibitem[Mueggler et~al.(2018)Mueggler, Gallego, Rebecq, and
  Scaramuzza]{Mueggler2018TRO}
E.~Mueggler, G.~Gallego, H.~Rebecq, and D.~Scaramuzza.
\newblock Continuous-time visual-inertial odometry for event cameras.
\newblock \emph{IEEE Transactions on Robotics (T-RO)}, 34\penalty0
  (6):\penalty0 1425--1440, 2018.

\bibitem[Park et~al.(2020)Park, Shin, and Kim]{Park2020ICRA}
Y.~S. Park, Y.~Shin, and A.~Kim.
\newblock Pharao: Direct radar odometry using phase correlation.
\newblock In \emph{IEEE International Conference on Robotics and Automation
  (ICRA)}, 2020.

\bibitem[Patron-Perez et~al.(2015)Patron-Perez, Lovegrove, and
  Sibley]{PatronPerez2015IJCV}
A.~Patron-Perez, S.~Lovegrove, and G.~Sibley.
\newblock A spline-based trajectory representation for sensor fusion and
  rolling shutter cameras.
\newblock \emph{International Journal of Computer Vision (IJCV)}, 113\penalty0
  (3):\penalty0 208--219, 2015.

\bibitem[Qin et~al.(2018)Qin, Li, and Shen]{Qin2018TRO}
T.~Qin, P.~Li, and S.~Shen.
\newblock Vins-mono: A robust and versatile monocular visual-inertial state
  estimator.
\newblock \emph{IEEE Transactions on Robotics (T-RO)}, 34\penalty0
  (4):\penalty0 1004--1020, 2018.

\bibitem[Shan et~al.(2020)Shan, Englot, Meyers, Wang, Ratti, and
  Daniela]{Shan2020IROS}
T.~Shan, B.~Englot, D.~Meyers, W.~Wang, C.~Ratti, and R.~Daniela.
\newblock Lio-sam: Tightly-coupled lidar inertial odometry via smoothing and
  mapping.
\newblock In \emph{IEEE/RSJ International Conference on Intelligent Robots and
  Systems (IROS)}, 2020.

\bibitem[Ye et~al.(2019)Ye, Chen, and Liu]{Ye2019ICRA}
H.~Ye, Y.~Chen, and M.~Liu.
\newblock Tightly coupled 3d lidar inertial odometry and mapping.
\newblock In \emph{IEEE International Conference on Robotics and Automation
  (ICRA)}, 2019.

\end{thebibliography}

\end{document}